\documentclass[a4paper]{article}
\usepackage{INTERSPEECH2019}
\usepackage[table,xcdraw]{xcolor}
\usepackage{amsmath}
\usepackage{amssymb}
\usepackage{dblfloatfix}
\usepackage{booktabs}
\usepackage{enumitem}
\usepackage{multirow}
\usepackage{tikz}
\usepackage{subcaption}
\usepackage{adjustbox}
\usepackage[singlelinecheck=false]{caption} %
\usepackage{textcomp}
\usepackage{units}
\usepackage[normalem]{ulem}
\usepackage{graphics}
\useunder{\uline}{\ul}{}
\usepackage[disable]{todonotes} %

\title{Objective Assessment of Social Skills Using Automated Language Analysis for Identification of Schizophrenia and Bipolar Disorder\thanks{Accepted to INTERSPEECH 2019 in Graz, Austria}}
\name{Rohit Voleti$^1$, Stephanie Woolridge$^3$, Julie M. Liss$^2$, Melissa Milanovic$^3$, Christopher R. Bowie$^3$, Visar Berisha$^{1,2}$}
\address{
  $^1$School of Electrical, Computer, \& Energy Engineering\\
  $^2$Department of Speech \& Hearing Science, Arizona State University, Tempe, AZ, USA\\ \vspace{0.15cm}
  $^3$Department of Psychology, Queen's University, Kingston, ON, Canada
}
\email{rnvoleti@asu.edu, 12smw2@queensu.ca, jmliss@asu.edu, melissa.milanovic@queensu.ca bowiec@queensu.ca, visar@asu.edu}

\newcommand{\etal}{\textit{et al}.\@ }
\newcommand{\ie}{\textit{i}.\textit{e}.\@ }
\newcommand{\eg}{\textit{e}.\textit{g}.\@ }
\newcommand{\etc}{\textit{etc}.\@}

\begin{document}
\maketitle

\begin{abstract}
Several studies have shown that speech and language features, automatically extracted from clinical interviews or spontaneous discourse, have diagnostic value for mental disorders such as schizophrenia and bipolar disorder.
They typically make use of a large feature set to train a classifier for distinguishing between two groups of interest, \ie a clinical and control group. 
However, a purely data-driven approach runs the risk of overfitting to a particular data set, especially when sample sizes are limited.
Here, we first down-select the set of language features to a small subset that is related to a well-validated test of functional ability, the Social Skills Performance Assessment (SSPA).
This helps establish the concurrent validity of the selected features. 
We use only these features to train a simple classifier to distinguish between groups of interest. 
Linear regression reveals that a subset of language features can effectively model the SSPA, with a correlation coefficient of {$0.75$}. 
Furthermore, the same feature set can be used to build a strong binary classifier to distinguish between healthy controls and a clinical group ($\text{AUC} = 0.96$) and also between patients within the clinical group with schizophrenia and bipolar I disorder ($\text{AUC} = 0.83$).
\end{abstract}
\section{Introduction \& Previous Work}\label{sec:intro}
In the United States alone, the National Institute of Mental Health (NIMH) in 2016 estimated that $\sim10.4$ million individuals live with a form of severe mental illness, approximately $4.2\%$ of the adult population \cite{center20172016}.
Among these are \emph{schizophrenia} and \emph{bipolar disorder}, for which diagnosis is difficult and treatment costs are disproportionately high \cite{desaiEstimatingDirectIndirect2013}.
Additionally, differential diagnosis of bipolar I disorder and schizophrenia is often difficult, with some estimating about $30$\% of bipolar I patients are misdiagnosed~\cite{altamuraDifferentialDiagnosesManagement2008}.
Therefore, there is a demand for effective methods with which we can classify and track the progress of treatment in these conditions.
Language impairments are a well-known component of schizophrenia and bipolar disorder, including symptoms like \emph{alogia} (poverty of speech) or development of \emph{formal thought disorder} (FTD), including \emph{schizophasia} ("word salad" or semantically incoherent utterances) \cite{APA2013}.
These impairments are typically assessed by clinical interviews, but few quantitative measures exist for measuring them objectively.
Recent work in computational linguistics and natural language processing (NLP) have paved the way for research into \emph{computational psychiatry} to objectively assess the degree of language impairment \cite{montagueComputationalPsychiatry2012}.
Several recent studies have made use of these tools for psychiatric evaluation, but their presence in clinical practice is still largely absent \cite{cecchiComputingStructureLanguage2017}.
In this paper, we aim to bridge this gap by presenting an objective and interpretable panel of language features for assessment of patients with schizophrenia and bipolar disorder that is anchored to a well-validated clinical assessment of social skills. 

Most existing work in this area takes a largely data-driven approach to language analysis, considering a host of semantic and lexical complexity measures over a large variety of language elicitation tasks \cite{elvevagQuantifyingIncoherenceSpeech2007, bediAutomatedAnalysisFree2015, corcoranPredictionPsychosisProtocols2018, iter2018automatic, W18-6211, motaThoughtDisorderMeasured2017}.
Semantic features are often captured with numerical word and sentence \emph{embeddings}, in which words, sentences, phrases, \etc are represented in high-dimensional vector space; typically, words that are semantically similar are embedded close together in this vector space, \eg \emph{latent semantic analysis} (LSA) \cite{landauerSolutionPlatoProblem1997}, \emph{word2vec} \cite{mikolovEfficientEstimationWord2013}, and several others.
Another measure of semantics can be achieved by topic modeling, such as with \emph{latent dirichlet allocation} (LDA) \cite{bleiLatentDirichletAllocation2003}.
Semantic features are often combined with other lexical measures of language complexity to improve classification performance. 
Some examples are ``surface features'' (\ie words per sentence, speaking rate, \etc) with tools like \emph{Linguistic Inquiry and Word Count} (LIWC) \cite{tausczik2010psychological}, statistical language features ($n$-gram word likelihoods) \cite{elvevagQuantifyingIncoherenceSpeech2007}, part-of-speech tag statistics \cite{bediAutomatedAnalysisFree2015, corcoranPredictionPsychosisProtocols2018, fraserLinguisticFeaturesIdentify2015}, and sentiment analysis \cite{socher2013recursive, kayiPredictiveLinguisticFeatures2017, mitchell2015quantifying}.

Despite promising early results, these tools are not currently used in clinical practice.
We posit that this is because the large and varied feature space, the variability associated with the speech elicitation tasks, and the small sample sizes make it difficult to develop reliable and interpretable algorithms that generalize.
As patient data is scarce, the identification of a standard set of important, interpretable, and easy-to-compute language features that clinicians can use is a significant hurdle to overcome. 
We address this by evaluating the language of patients with schizophrenia, patients with bipolar I disorder, and healthy control subjects on the \emph{Social Skills Performance Assessment} (SSPA) \cite{pattersonSocialSkillsPerformance2001}, a well-validated test of social functional competence (described in Section~\ref{sec:sspa}).
Our approach is motivated by our previous work in interpretable clinical-speech analytics \cite{tuInterpretableObjectiveAssessment2017}.
First, we identify a subset of language measures that reliably model clinical SSPA scores.
Next, we use only this reduced feature set to perform two classification problems: ($1$) distinguishing between healthy controls and clinical subjects and ($2$) distinguishing patients with schizophrenia/schizoaffective disorder (Sz/Sza) and bipolar I patients within the clinical group.
To the best of our knowledge, this is the first study to establish a set of language measures that jointly assess social skills \emph{and} uses those features to accurately classify all groups of interest. %
\section{SSPA Data Collection}\label{sec:sspa}
Our study involves the analysis of interview transcripts collected from a total of $87$ clinical subjects and $22$ healthy controls that participated in the SSPA task described by Patterson \etal \cite{pattersonSocialSkillsPerformance2001}.
Of the clinical population, $44$ had been diagnosed with bipolar I disorder and $43$ had been diagnosed with schizophrenia or schizoaffective disorder (considered together in this analysis).
The SSPA interviews are described by Bowie \etal in \cite{bowiePredictionRealworldFunctional2010}. 
The transcriptions used in our analysis were completed at Queen's University in Kingston, ON, Canada.

The task consists of three role-playing scenes: ($1$) $1$-minute practice scene of making plans with a friend (not scored), ($2$) $3$ minutes of greeting a new neighbor, and ($3$) $3$ minutes of negotiation with a recalcitrant landlord over fixing an unrepaired leak.
Each session was recorded and scored by trained research assistants upon reviewing the recording.
Scene $2$ (new neighbor) and Scene $3$ (negotiation with landlord) were scored on a scale of $1$ (low) to $5$ (high) on several categories, \ie interest/disinterest, fluency, clarity, social appropriateness, negotiation ability, \etc~
A composite score for each scene and an overall score is computed by averaging Scene $2$ and Scene $3$ scores.

Bowie \etal identified group differences between the scores of both clinical populations and healthy control subjects in \cite{bowiePredictionRealworldFunctional2010} by evaluation on the SSPA task and several other clinical measures.
In this work, we aim to automate this task with a subset of language metrics from the SSPA transcripts.
As stated in Section \ref{sec:intro}, our first goal is to identify semantic and lexical features from which we can reliably predict SSPA performance.
Then, we test the ability of these features to differentiate between healthy control and clinical populations, and we also test their ability to differentiate within the distinct groups in the clinical population.
\section{Computed Language Features}\label{sec:features}
In our work, we attempt to identify a comprehensive set of objective language measures from which we can model and predict SSPA performance and classify individuals using these features.
Inspired by much of the previous work described in Section \ref{sec:intro}, we theorized that it is critical to consider language features that model semantic coherence through the use of word and sentence embeddings.
We focused on a few pre-trained neural embedding models that are publicly available and known to model semantic similarity accurately.
Additionally, we consider a set of lexical complexity features that are measures of lexical and syntactic complexity, described below.

\subsection{Semantic Coherence}
Many of the previously described studies in this area involve computing a notion of semantic coherence in language with the use of word embeddings in high-dimensional vector space, either with LSA or neural word embedding techniques \cite{elvevagQuantifyingIncoherenceSpeech2007, bediAutomatedAnalysisFree2015, corcoranPredictionPsychosisProtocols2018, iter2018automatic}.
In nearly all cases, word and sentence/phrase embedding pairs, denoted by vectors $\mathbf{a}$ and $\mathbf{b}$, are evaluated with the notion of \emph{cosine similarity}, a measure of the cosine of the angle $\theta_{\mathbf{a}, \mathbf{b}}$ between the two vectors. We also use cosine similarity as a measure of pairwise sentence similarity, but with some modifications in implementation due the difference in the nature of the SSPA task and data collection.

Our work differs from several of the previously discussed studies in that we are interested in conversational semantic similarity between the subject and clinical assessor in each of the three scenes of the SSPA task. %
Therefore, we sought to utilize some of the latest sentence/phrase embedding methods to compute a vector representation for each assessor and subject speaking turn.
Then, we used the cosine similarity to compute the similarity score between each consecutive assessor $+$ subject speaking turn, generating a distribution of similarity scores for each embedding method for each subject in each transcribed scene. 
The following sentence embedding representations are used in our analysis: ($1$) an unweighted bag-of-words (BoW) average for all word vectors based on the pre-trained \emph{skip-gram} implementation of \emph{word2vec} trained on the Google News corpus \cite{mikolovEfficientEstimationWord2013}, ($2$) \emph{Smooth Inverse Frequency} (SIF) with pre-trained skip-gram \emph{word2vec} vectors \cite{aroraSimpleToughtoBeatBaseline2017}, and ($3$) \emph{InferSent} (INF) sentence encodings based on pre-trained \emph{FastText} vectors \cite{conneauSupervisedLearningUniversal2017}. 
The BoW average of vectors and SIF embeddings showed good baseline performance in \cite{iter2018automatic}, and we additionally included \emph{InferSent}, a deep neural network sentence encoder, due to its strong performance on semantic similarity tasks.
Then, basic statistics for the similarity score distribution were computed for each subject and transcribed scene.
These included minimum, maximum, mean, median, $90^{\text{th}}$ percentile, and $10^{\text{th}}$ percentile coherence.

\subsection{Linguistic Complexity}
While semantic coherence measures are often the most effective at classifying patients with schizophrenia and bipolar disorder, several other linguistic complexity measures are used for a more holistic analysis.
We consider a subset of these features, computed for the entire set of subject responses across all three scene transcripts.

\emph{Lexical diversity} refers to unique vocabulary usage for a particular subject and for which several measurement techniques exist.
The \emph{type-to-token ratio} (TTR) is a well-known measure of lexical diversity, in which the number of unique words (word \emph{types}, $V$) are compared against the total number of words (word \emph{tokens}, $N$): $\mathrm{TTR} = \nicefrac{V}{N}$.
However, TTR is known to be negatively impacted for longer utterances, as the diversity of unique words plateaus as the number of total words increase.
Hence, we consider a small selection of modified measures for lexical diversity in our work.
The moving average type-to-token ratio (MATTR) \cite{covingtonCuttingGordianKnot2010} is one such method which aims to reduce the dependence on text length by considering TTR over a sliding window of the text.
\emph{Brun\'et's Index} (BI) \cite{brunet1978vocabulaire}, defined in Equation~(\ref{eq:brunet}), is another measure of lexical diversity that has a weaker dependence on text length.
A smaller value indicates a greater degree of lexical diversity
\begin{equation}\label{eq:brunet}
	\mathrm{BI} = N^{V^{-0.165}}
\end{equation}
An alternative is also provided by \emph{Honor\'e's Statistic} (HS) \cite{honore1979some}, defined in Equation~(\ref{eq:honore}), which emphasizes the use of words that are spoken only once (denoted by $V_1$).
\begin{equation}\label{eq:honore}
	\mathrm{HS} = 100 \log \frac{N}{1 - \nicefrac{V_1}{V}}
\end{equation}
MATTR, BI, and HS have been used successfully in computational linguistics studies for patients with Alzheimer's disease \cite{ fraserLinguisticFeaturesIdentify2015, bucksAnalysisSpontaneousConversational2000} and may prove to be similarly useful in our task.

Because we expect schizophrenia and bipolar patients to sometimes exhibit poverty of speech, we considered a few measures of lexical and syntactic complexity in our work. 

\emph{Lexical density}, which quantifies the degree of information packaging in a given text, is defined as the proportion of \emph{content words} (\ie nouns, verbs, adjectives, adverbs) \cite{johansson2009lexical}.
Typically, these words convey more information than \emph{function words}, \eg prepositions, conjunctions, interjections, \etc~
We make use of the Stanford tagger \cite{toutanovaFeaturerichPartofspeechTagging2003} to compute POS tags to determine the number of function words (FUNC) and total words (W) and measure $\nicefrac{\text{FUNC}}{\text{W}}$, which represents an inverse of the lexical density.
A related, more granular measure is the proportion of interjections (UH) to the total words, which is given by $\nicefrac{\text{UH}}{\text{W}}$.
The mean length of sentence (MLS) is another easily computed measure which we expect to be lower for clinical subjects when compared with healthy controls.
Finally, we considered parse tree statistics, computed using the Stanford Parser \cite{Socher13parsingwith}.
This includes the parse tree height and Yngve depth scores (mean, total, and maximum), a measure of embedded clause usage \cite{yngveModelHypothesisLanguage1960}.
\begin{figure}
    \centering
    \includegraphics[width=\linewidth]{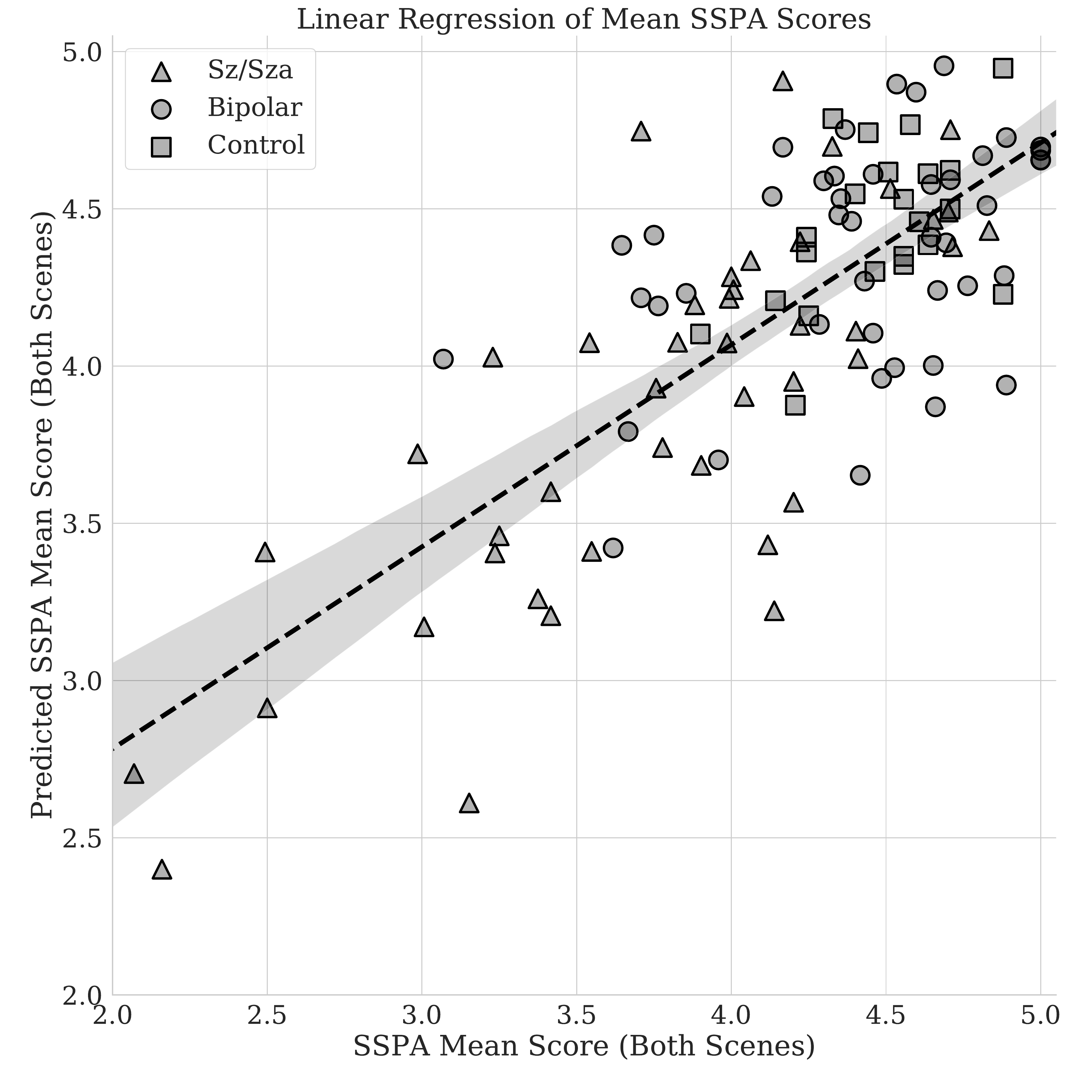}
    \caption{A linear regression model was fit using $25$ out of the $73$ semantic coherence and linguistic complexity features from the $109$ subject responses to predict the SSPA scores. Correlation Coefficient = $\mathbf{0.752}$, Mean Absolute Error = $\mathbf{0.330}$, Root Mean Square (RMS) Error = $\mathbf{0.405}$}
    \label{fig:sspa_reg}
    \vspace{-0.3cm}
\end{figure}

\vspace{-0.3cm}
\section{Results \& Discussion}

We first sought to determine a subset of language features (described in Section \ref{sec:features}) from which we can accurately model the clinical SSPA scores.
A total of $73$ features were considered: $63$ semantic features ($7$ statistical features $\times$ $3$ sentence embedding types $\times$ $3$ scenes) and $10$ linguistic complexity features computed over all three scenes concatenated.
Next, we aim to determine the predictive power of the selected subset of these features in separating the groups of interest (\ie Sz/Sza, bipolar I disorder, and healthy control subjects).
The regression and classification models built with these features were designed and tested using WEKA  \cite{hallWEKADataMining2009}.
It is important to note that the SSPA itself is correlated to the clinical diagnosis and has been effective in differentiating groups of interest~\cite{bowiePredictionRealworldFunctional2010}.
As a result, we note that using it to select features may result in overly-optimistic classification performance for the clinical vs. healthy control and Sz/Sza vs. bipolar disorder classification problems. 
However, due to the relative dearth of available data in this area, we performed this analysis on the same dataset.

\begin{table}[ht!]
    \centering
    \footnotesize
	\caption{Selected features to model SSPA scores with a linear regression model, including ranking of overall importance for each feature. Italicized features were included in both the 25 feature and 15 feature classification problems.}
	\label{table:features}
	\begin{tabular*}{0.9\columnwidth}{@{}llr@{}}
		\toprule
		Category & Features & Rank \\ \midrule
		Semantic Coherence & \textit{BoW mean scene 3} & 1 \\
		& \textit{INF minimum scene 3} & 2 \\
		& \textit{SIF 90$^{\text{th}}$ percentile scene 3} & 5 \\
		& \textit{INF maximum scene 2} & 7 \\
		& \textit{INF median scene 3} & 8 \\
		& \textit{BoW median scene 3} & 9 \\
		& \textit{BoW minimum scene 2} & 10 \\
		& \textit{BoW st. dev. scene 2} & 11 \\
		& \textit{BoW maximum scene 3} & 12 \\
		& \textit{INF st. dev. scene 3} & 13 \\
		& BoW maximum scene 2 & 18 \\
		& BoW 90$^{\text{th}}$ percentile scene 2 & 19 \\
		& BoW st. dev. scene 3 & 20 \\
		& BoW 90$^{\text{th}}$ percentile scene 3 & 21 \\
		& INF mean scene 3 & 22 \\
		& INF 10$^{\text{th}}$ percentile scene 3 & 23 \\
		& BoW 10$^{\text{th}}$ percentile scene 2 & 24 \\
		Lexical Diversity & \textit{MATTR} & 3 \\
		& \textit{Brun\'et's index} & 4 \\
		& Honor\'e's statistic & 25 \\
		Lexical Density & \textit{FUNC/W} & 6 \\
		& \textit{UH/W} & 14 \\
		Syntactic Complexity & \textit{Maximum Yngve depth} & 15 \\
		& Mean length sent. (MLS) & 16 \\
		& Parse tree height & 17 \\ \bottomrule
	\end{tabular*}
\end{table}
\begin{figure}[t]
    \centering
    \includegraphics[width=\linewidth]{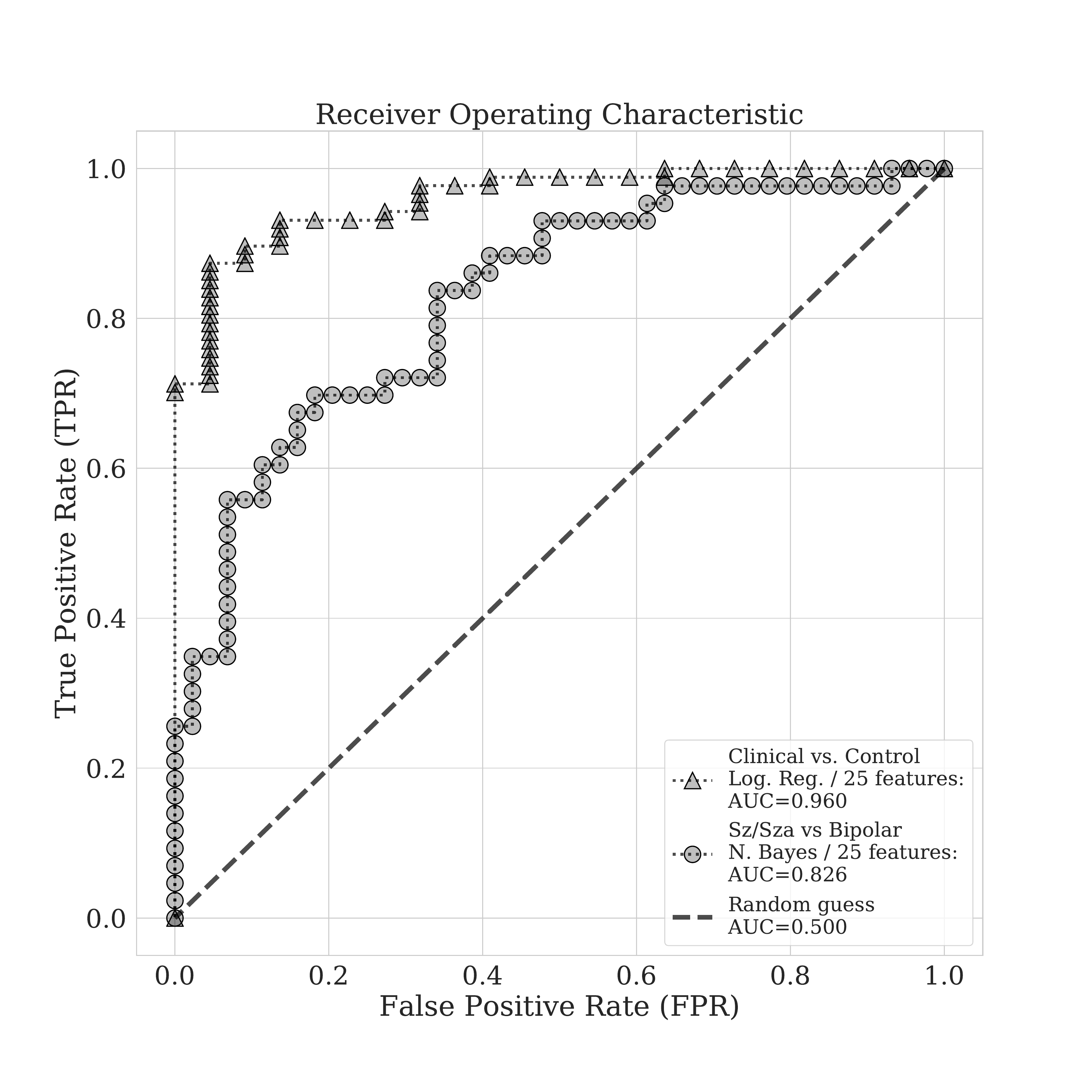}
    \caption{Selected receiver operating characteristic (ROC) curves for both binary classification tasks. For clinical vs control classification, TPR indicates correctly classifying a clinical subject and FPR indicates falsely classifying a control subject as clinical. For Sz/Sza vs bipolar classification, TPR is correctly classifying an Sz/Sza subject and FPR is falsely classifying a bipolar subject as Sz/Sza.}
    \label{fig:roc}
\end{figure}
\begin{table}[]
    \footnotesize
	\caption{Confusion matrices for binary classification results with logistic regression (LR) and na\"ive Bayes(NB) classifiers with a $25$ feature and $15$ feature subset. (\subref{table:clin-cont-class}) For clinical vs control classification, LR with $25$ features works best at differentiating groups. (\subref{table:schiz-bipolar-class}) For Sz/Sza vs bipolar classification, LR using a $25$ feature subset works poorly. NB provides more consistent results, even when the feature set is reduced.}
	\label{table:confusion}
	\resizebox{.85\columnwidth}{!}{
	\begin{subtable}{.73\linewidth}
		\caption{Clinical vs Control}
		\label{table:clin-cont-class}
		{\begin{tabular}{@{}lllrr@{}}
				\toprule
				\textit{} & \textit{} &  & \multicolumn{2}{l}{True group:} \\
				\multicolumn{2}{l}{\textit{Log. Reg.}} &  & Clinical & Control \\ \cmidrule(l){4-5} 
				& \multicolumn{1}{r}{\multirow{2}{*}{25 feat.}} & \multicolumn{1}{r|}{Clinical} & 78 & 3 \\
				& \multicolumn{1}{r}{} & \multicolumn{1}{r|}{Control} & 9 & 19 \\
				&  &  & \multicolumn{2}{l}{AUC =  0.960} \\
				&  &  & \multicolumn{1}{l}{} & \multicolumn{1}{l}{} \\
				&  &  & Clinical & Control \\ \cmidrule(l){4-5} 
				& \multicolumn{1}{r}{\multirow{2}{*}{15 feat.}} & \multicolumn{1}{r|}{Clinical} & 79 & 10 \\
				& \multicolumn{1}{r}{} & \multicolumn{1}{r|}{Control} & 8 & 12 \\
				&  &  & \multicolumn{2}{l}{AUC = 0.882} \\
				&  &  & \multicolumn{1}{l}{} & \multicolumn{1}{l}{} \\
				\multicolumn{2}{l}{\textit{N. Bayes}} &  & Clinical & Control \\ \cmidrule(l){4-5} 
				& \multicolumn{1}{r}{\multirow{2}{*}{25 feat.}} & \multicolumn{1}{r|}{Clinical} & 73 & 2 \\
				& \multicolumn{1}{r}{} & \multicolumn{1}{r|}{Control} & 14 & 20 \\
				&  &  & \multicolumn{2}{l}{AUC = 0.908} \\
				&  &  & \multicolumn{1}{l}{} & \multicolumn{1}{l}{} \\
				& \multicolumn{1}{r}{} & \multicolumn{1}{r}{} & Clinical & Control \\ \cmidrule(l){4-5} 
				& \multicolumn{1}{r}{\multirow{2}{*}{15 feat.}} & \multicolumn{1}{r|}{Clinical} & 76 & 5 \\
				& \multicolumn{1}{r}{} & \multicolumn{1}{r|}{Control} & 11 & 17 \\
				&  &  & \multicolumn{2}{l}{AUC = 0.873} \\ \bottomrule
		\end{tabular}}
	\end{subtable}
	\begin{subtable}{.27\linewidth}
		\caption{Sz/Sza vs Bip.}
		\label{table:schiz-bipolar-class}
		{\begin{tabular}{@{}lrr@{}}
				\toprule
				& \multicolumn{2}{l}{True group:} \\
				& Sz/Sza & Bipolar \\ \cmidrule(l){2-3} 
				\multicolumn{1}{r|}{Sz/Sza} & 30 & 14 \\
				\multicolumn{1}{r|}{Bipolar} & 13 & 30 \\
				& \multicolumn{2}{l}{AUC =  0.700} \\
				& \multicolumn{1}{l}{} & \multicolumn{1}{l}{} \\
				& Sz/Sza & Bipolar \\ \cmidrule(l){2-3} 
				\multicolumn{1}{r|}{Sz/Sza} & 30 & 10 \\
				\multicolumn{1}{r|}{Bipolar} & 13 & 34 \\
				& \multicolumn{2}{l}{AUC = 0.796} \\
				& \multicolumn{1}{l}{} & \multicolumn{1}{l}{} \\
				& Sz/Sza & Bipolar \\ \cmidrule(l){2-3} 
				\multicolumn{1}{r|}{Sz/Sza} & 30 & 11 \\
				\multicolumn{1}{r|}{Bipolar} & 13 & 33 \\
				& \multicolumn{2}{l}{AUC = 0.826} \\
				& \multicolumn{1}{l}{} & \multicolumn{1}{l}{} \\
				\multicolumn{1}{r}{} & Sz/Sza & Bipolar \\ \cmidrule(l){2-3} 
				\multicolumn{1}{r|}{Sz/Sza} & 31 & 11 \\
				\multicolumn{1}{r|}{Bipolar} & 12 & 33 \\
				& \multicolumn{2}{l}{AUC = 0.803} \\ \bottomrule
		\end{tabular}}
	\end{subtable}
	}	
\end{table}

\subsection{Modeling SSPA Performance}\label{subsec:sspa}
We use a greedy stepwise search (with linear regression) through the feature space to determine the optimal subset of the features which accurately model the SSPA scores for all $109$ subjects without considering the group variable. We down-selected to a set of $25$ computed features out of the original $73$.
These are briefly summarized in Table~\ref{table:features}, and the resulting regression model (evaluated using leave-one-out) is shown in Figure~\ref{fig:sspa_reg}.
We notice that several of the coherence statistics for Scene $3$ (negotiation with landlord) are particularly influential when tracking the assigned SSPA score with this model.
Interestingly, the top three coherence statistics include a bag-of-words average of \emph{word2vec} vectors (BoW mean scene $3$), an \emph{InferSent} sentence encoding (INF minimum scene $3$), and a SIF embedding (SIF $90^{\text{th}}$ percentile scene $3$), indicating a variety of embeddings and range of statistics all provide useful information in predicting SSPA performance.
We also note that a variety of lexical diversity (MATTR, Brun\'et's index), lexical density ($\nicefrac{\text{FUNC}}{\text{W}}$, $\nicefrac{\text{UH}}{\text{W}}$) and syntactic complexity (maximum Yngve depth) measures are among the most influential, confirming the benefit of a complementary set of language measures.

\subsection{Identification of Schizophrenia and Bipolar Disorder}\label{subsec:class}
Next, we aim to determine the ability of this subset of language features to correctly predict which subjects fall into the groups of interest.
We performed two separate classification tasks: ($1$) separation of the clinical and healthy control groups, ($2$) separation within the clinical group between Sz/Sza subjects and bipolar I subjects.
Both a logistic regression (LR) and a na\"ive Bayes (NB) classifier were trained in each case using leave-one-out cross validation to determine model parameters and performance.
Then, we further down-selected this set to a group of $15$ features and re-evaluated the performance of both classifiers.

The confusion matrices for the clinical and control group classification task are shown in Table~\ref{table:clin-cont-class}.
As we can see, LR with all $25$ selected features works best, with the area under curve (AUC) in the ROC plot being $0.960$ (see Figure~\ref{fig:roc}).
In this case, $78$ of $87$ ($89.7\%$) clinical subjects and $19$ of $22$ ($86.7\%$) healthy controls were correctly identified in our leave-one-out evaluation.
We also see comparable performance for the NB and LR models when the feature set is reduced to only the top $15$ features that model SSPA scores, though AUC is lower than both models with $25$ features.

Next, we consider a classification problem within the group of $87$ clinical subjects, of which $43$ are diagnosed with Sz/Sza and 44 are diagnosed with bipolar I disorder.
We use the same feature subsets and same binary classifier models as in the previous task, trained and evaluated using leave-one-out cross-validation.
From the confusion matrices in Table~\ref{table:schiz-bipolar-class}, we see that NB performs better than LR when either a $25$ feature or $15$ feature subset are used, with the best $\text{AUC} = 0.826$ for NB with 25 features.
The ROC curve for a $25$-feature NB classifier is shown in Figure~\ref{fig:roc}.
Interestingly, LR with $25$ features had the lowest performance on this task ($\text{AUC} = 0.700$).

LR typically performs better than NB when more data is available for training \cite{ng2002discriminative}; however in clinical applications data set size is often limited.
This makes sense with respect to our study, as the dataset used in the Sz/Sza vs. bipolar I classification problem is smaller than the dataset used in the clinical vs control group classification problem.  
In this case, the LR model is prone to overfitting, as is evident by the fact that performance improves when the feature dimension is reduced. 
As expected, the classifier performance is considerably worse than the clinical and control group classification problems, as the language differences between schizophrenia and bipolar patients are more difficult to distinguish, even for experienced clinicians.
Considering this fact, we still see reasonable performance with only computed language measures and no additional clinical assessment.
\section{Conclusion}
This paper demonstrates the potential of computational linguistics to aid neuropsychiatric practice in the clinic.
We believe it is critically important to tie computational methods to established clinical practice in order to bridge the gap between the latest developments in NLP, which motivated our feature selection using SSPA.
Still, there are many directions in which we can take future work.
The sentence embedding and coherence metrics computed in this study are by no means an exhaustive list of potential methods, and it is likely a more optimal easily computable feature set exists to model SSPA performance and classify groups of interest.
In particular, we are interested in finding a more concise group of clinically relevant language features with which we can perform this analysis.
Additionally, we can look at more language metrics within each subject group to further subtype and cluster individuals within each group based on language metrics.
These methods can also be applied to clinical assessments beyond the SSPA tasks and for a wider variety of psychiatric conditions.
Lastly, we would like to examine how classification and modeling of clinical test scores changes when computed features are used in conjunction with other clinical tests to model task performance and classification of groups.
\section{Acknowledgment}

This work was partially funded by a grant from the Boeheringer Ingelheim International GmbH to ASU (PI: Berisha). %

\bibliographystyle{IEEEtran}

\bibliography{references/refs}

\end{document}